\documentclass[conference]{IEEEtran}
\IEEEoverridecommandlockouts
\usepackage{cite}
\usepackage{amsmath,amssymb,amsfonts}
\usepackage{algorithmic}
\usepackage{graphicx}
\usepackage{textcomp}
\usepackage{xcolor}
\def\BibTeX{{\rm B\kern-.05em{\sc i\kern-.025em b}\kern-.08em
    T\kern-.1667em\lower.7ex\hbox{E}\kern-.125emX}}

\usepackage[hidelinks]{hyperref}

\usepackage{adjustbox}

\begin{document}

\title{A novel network for classification of cuneiform tablet metadata
\thanks{This project was funded by Innovation Fund Denmark through the projects MADE ReAct and FERA, and by Fabrikant Vilhelm Pedersen og Hustrus Legat.}
}

\author{\IEEEauthorblockN{1\textsuperscript{st} Frederik Hagelskjær*}
\IEEEauthorblockA{\textit{SDU Robotics} \\
\textit{The Mærsk Mc-Kinney Møller Institute} \\
\textit{University of Southern Denmark}\\
Odense, Denmark \\
frhag@mmmi.sdu.dk}
*Corresponding author
}

\maketitle


\begin{abstract}
In this paper, we present a network structure for classifying metadata of cuneiform tablets. The problem is of practical importance, as the size of the existing corpus far exceeds the number of experts available to analyze it. But the task is made difficult by the combination of limited annotated datasets and the high‑resolution point‑cloud representation of each tablet. To address this, we develop a convolution‑inspired architecture that gradually down-scales the point cloud while integrating local neighbor information. The final down-scaled point cloud is then processed by computing neighbors in the feature space to include global information. 
Our method is compared with the state-of-the-art transformer-based network Point-BERT, and consistently obtains the best performance. Source code and data available at \url{github.com/fhagelskjaer/cuneiform3d}
\end{abstract}


\section{Introduction}

\begin{figure}[t]
    \centering
    \includegraphics[width=0.99\linewidth]{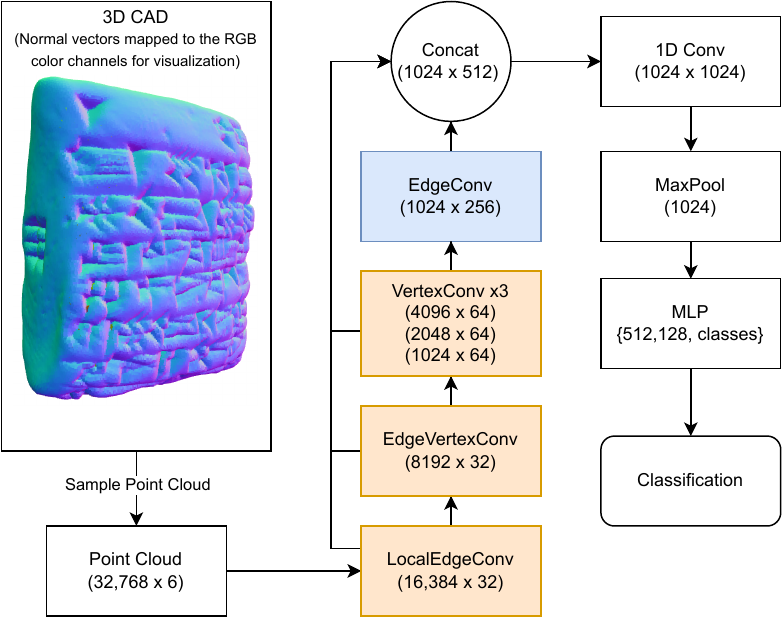}
    \caption{The network structure of our presented method. First a point cloud is sampled from the CAD model. The point cloud is then processed and down-sampled. The orange layers compute neighbors in spatial space. In the blue layer neighbors are found the feature space. The tablet is "HS 2274" \cite{mara2019breaking}.
    }
    \label{fig:enter-label}
\end{figure}

The production of cuneiform tablets spanned the period from the fourth to the first millennium BCE \cite{walker1987cuneiform}. As cuneiform is written by imprinting wedges onto clay tablets, it has a high durability compared to other media for recording. The durability and long production period, has led to the excavation of hundreds of thousands of tablets. While some tablets provide substantial knowledge independently (e.g., the Complaint tablet to Ea-nāṣir \cite{moshenska2023museum}), most tablets contain information that, on its own, provides little understanding. However, a combined corpus of tablets can enable a much deeper understanding of the period \cite{page2018network}. 
Despite this potential, many tablets have not yet been analyzed, as the number of experts in cuneiform is small relative to the volume of material \cite{cohen2004iclay}. 
To address this, deep learning-based methods have been proposed as a solution, and have shown promising results \cite{kapon2024shaping}. Most existing approaches process the tablets as 2D images, since a wide range of mature methods already exists for this modality \cite{stotzner2023cnn}. However, cuneiform tablets are inherently 3D objects with text often wrapping around the corners \cite[p.~14]{walker1987cuneiform}. As such, information can be lost when tablets are flattened to 2D images. An example of successful 3D processing was demonstrated by Bogacz et al. \cite{bogacz2020period}, by predicting the time period of tablets, with results further improved in \cite{hagelskjaer2022deep}, which also introduced two additional classification tasks, seal presence and left-side sign detection. However, as the datasets are very small, ranging from 337 to 747 tablets, the risk of overfitting is very large.
One approach to overcome the challenge of limited training data is to use smaller receptive fields. As such CNN-like (Convolutional Neural Network) structures demonstrate better performance than transformer based models when limited training data is available \cite{mauricio2023comparing}. We build on an existing method \cite{hagelskjaer2025arrowpose} to build a multilayered architecture for point clouds with a CNN-like structure that gradually increases the receptive field, by combining PointNet++ \cite{qi2017pointnet++} down-sampling and DGCNN neighbor features \cite{dgcnn}. We improve the structure by introducing new neighbor features, and by integrating the feature space neighbor layer after down-sampling the point cloud.
The network structure is shown in Fig.~\ref{fig:enter-label}. 
A different approach to avoid overfitting on small datasets is to use generalized models pre-trained on large amount of data. The weights are then kept frozen while a small classification head is trained for the task \cite{xue2024ulip}. 
Point-BERT \cite{yu2022point} pre-trained using ULIP-2 \cite{xue2024ulip} is a notable example for point clouds and have shown state-of-the-art performance for zero-shot classification tasks. We compare our method with Point-BERT on the datasets from \cite{bogacz2020period} and \cite{hagelskjaer2022deep}, and while Point-BERT achieves performance comparable to \cite{hagelskjaer2022deep}, our network show even better performance on all datasets. 
%
%
Additionally, we introduce a new classification task of determining the front of the tablet, a difficult task \cite{mara2019breaking}, which is best fit for 3D data \cite[p.~22]{walker1987cuneiform}. On this task our network shows very good performance and actually detects an error in the dataset.
These results demonstrate that, for tasks with limited training data, a more structured network can outperform transformer-based models.

\section{Related Work}



In this section we give an overview of the current state-of-the-art in point cloud classification, starting with the PointNet architecture PointNet \cite{qi2017pointnet}.
%
%
To address the unordered structure of point cloud data, PointNet used a global MaxPool across all points after computing features using deep learning. 
This feature from the MaxPool layer is then processed by an MLP (Multi Layer Perceptron) to provide classification predictions. PointNet demonstrated state-of-the-art performance for both classification and segmentation in point clouds. 
This MaxPool is universal for all subsequent development in deep learning of point clouds, and we also apply this for our network.
A weakness of PointNet is that the features of each point is computed completely independently. Thus it is not able to incorporate local neighborhood information into the feature. A number of methods have improved on the performance of PointNet by allowing points to encode information from other points in the point cloud. 
PointNet++ \cite{qi2017pointnet++} uses a sub-sampling strategy to first compute smaller PointNet features for a number of keypoints. These features are then used as input for a second PointNet. By gradually performing this operation, features are aggregated and the receptive field of each PointNet is increased. This approach demonstrated increased performance compared with PointNet.
%
%
DGCNN \cite{dgcnn} includes neighbor information in a different approach. For all points a k-NN is computed. A 1D-convolution is then applied on each point-neighbor pair. Finally, a local MaxPool computes the strongest feature response of the point-neighbor pairs. This feature is computed for each point. In the subsequent layers the neighbor are found in the feature space, this sets the receptive field to the full point cloud. Compared with PointNet++ the local neighborhoods are much smaller, the largest in PointNet++ is 512 where compared with twenty in DGCNN. As the receptive field is increased by finding neighbors in feature space the down-sampling is not required.
ArrowPose \cite{hagelskjaer2025arrowpose} presented a method which combined the sub-sampling of PointNet++ with the local neighborhood feature from DGCNN. This network structure allows for processing very large point clouds allows, and is used for segmentation. We use this structure to allow for the very large point clouds from the tablets. 
Dilation is an alternative to down-sampling for increasing the receptive field. By dilation a number of points are skipped during the neighbor search. This allows for a further receptive field without sub-sampling. Dilated Point Convolutions \cite{engelmann2020dilated} demonstrated improved performance by using this method. We also employ dilation in our method and show improved performance.
%
Another approach to processing point clouds is by conversion to voxel grids, the 3D equivalent of the 2D CNN feature grid. A drawback is that the resolution grows cubically with grid size, forcing the grid to remain relatively coarse. In VoxelNet \cite{zhou2018voxelnet}, the point cloud is split into voxels, and a PointNet feature is computed for each voxel and 3D-convolutions are then applied. This hybrid approach has shown very good results, as it allows for more fine-grained computations, but it still suffers from the coarse resolution.


Transformer based models where initially introduced for natural language processing, but have shown impressive results for 2D processing compared with CNN based models \cite{mauricio2023comparing}. This trend has also developed into 3D processing. Point-BERT \cite{yu2022point}, is a notable example which has outperformed previous CNN based methods such as PointNet++ and DGCNN. Similar to PointNet++ the first operation of Point-BERT is a computation of keypoints, sub-sampling and computing small PointNet features. These features are then provided to a transformer encoder. After processing of multiple layers a final CLS (classification) token is given to an MLP for classification.
To overcome the requirement for large amounts of training data that transformer based models generally require, Point-BERT introduces a pre-training step. In this pre-training step, the model learns to infer the geometric structure of missing parts of the point cloud. This enables the model to learn stronger feature representations using a smaller amount of training data.
An alternative to training for a specific task is to train on general datasets to create foundation models. Using either fine-tuning or zero-shot models from these have shown impressive results on a realm of different tasks. This has also been demonstrated for point cloud data. In ULIP-2 \cite{xue2024ulip} a Point-BERT network is trained to align the computed 3D-features with features of the same object from image and text data. The images are generated by synthetic rendering and the text data is generated automatically using BLIP-2 \cite{li2023blip} and then filtered using CLIP \cite{radford2021learning}. This self-supervised approach allow for generating large amounts of training data without a requirement for manual annotation. Using this approach obtains state-of-the-art results for zero-shot classification of 3D. In this paper we show results for Point-BERT pre-trained on the ULIP-2 dataset. The ULIP-2 weights are kept frozen and we then fine-tune a DGCNN classification head.

\section{Method}

The main contribution of this paper is a novel network for classifying large point clouds. The network combines the down-scaling of PointNet++ \cite{qi2017pointnet++} with the feature space neighbors from DGCNN \cite{dgcnn}. In the following section, we elaborate on the network structure and the training. Finally, we explain the fine-tuning of Point-BERT.

\subsection{Network Structure}

The base structure of our network is inspired by ArrowPose \cite{hagelskjaer2025arrowpose}, which integrates the neighborhood feature from DGCNN \cite{dgcnn} with the down-scaling presented in PointNet++ \cite{qi2017pointnet++}. This creates a structure similar to that of classical CNNs, with neighbor computation, pooling, and down-sampling. After each step, the point cloud size is halved, increasing the receptive field. Down-sampling is performed by randomly shuffling the point cloud and truncating it; yielding a stochastically even representation. Random down-sampling have shown to provide similar results as Farthest Point Sampling while being much faster \cite{hu2020randla}.
In ArrowPose, the final layer of points is used to predict segmentations and object-center locations for pose estimation. In this paper, we instead use the framework for classification. We replace the segmentation part from ArrowPose with a MaxPool and the MLP classification head from DGCNN. Additionally, to improve performance, we introduce several additions to the network. After the network computation and down-sampling, we use the EdgeConv neighbor features from DGCNN. This computes a feature space neighbor and provides a global receptive field in the final layer. We then concatenate the features from all layers and apply a 1D convolutional network. This network is applied before the MaxPool layer to aggregate information across layers into a single feature. This helps to ensure that all features representing a single point are combined before the MaxPool layer. We also use dilation during the neighbor search, similar to \cite{engelmann2020dilated}, which shows to improve the results. Finally, we introduce two new neighbor features, which are further explained in the following subsection.









\subsection{Neighbor Features}


The neighbor feature used in DGCNN \cite{dgcnn} is the EdgeConv operator (Eq. (1) in \cite{dgcnn}). For each point, DGCNN identifies its $n$ nearest neighbors in feature space and computes a pairwise feature by concatenating the point feature with its relative difference to each neighbor. A shared 1D convolution is then applied:
\begin{equation}
\bar{h}_{\theta}(x_i, x_j - x_i)
\label{eq:ec}
\end{equation}
where $x_i$ is the point feature, $x_j$ is the neighbor feature, and $\bar{h}_{\theta}$ denotes the learned weights. The final EdgeConv output is obtained by applying MaxPool over all neighbors:
\begin{equation}
\text{MaxPool}\left(\bar{h}{\theta}(f_i,f_1-f_i),\Vert,\cdots,\Vert,\bar{h}{\theta}(f_i,f_n-f_i)\right)
\label{eq:maxec}
\end{equation}

\begin{table}[t]
    \centering
    \caption{Hyper parameters for each spatial neighbor layer.}
    \label{tab:network_values}
    \begin{adjustbox}{width=0.49\textwidth}
    \begin{tabular}{lcccccc}
        \hline 
        Layer & 1 & 2 & 3 & 4 & 5 \\ 
        \hline
        Conv type & LocalEdge & EdgeVertex & Vertex & Vertex & Vertex \\
        Input size & 32,768 & 16,384 & 8192 & 4096 & 2048 \\ 
        Output size & 16,384 & 8192 & 4096 & 2048 & 1024 \\ 
        Features & 32 & 32 & 64 & 64 & 64 \\ 
        Dilation & 1 & 1 & 2 & 2 & 1 \\ 
        Neighbors & 16 & 16 & 16 & 16 & 16
    \end{tabular}
    \end{adjustbox}
\end{table}

Because DGCNN computes neighbors in feature space, the k-NN search must be repeated after each layer. Performing this search on the GPU requires constructing an $N \times N$ distance matrix, which scales quadratically with the point cloud size and becomes prohibitive for large point clouds \cite{hagelskjaer2022deep}.
To enable processing of very large point clouds, ArrowPose instead computes neighbors solely based on spatial distance, following the strategy of PointNet++ \cite{qi2017pointnet++}. Spatial neighbors can be pre-computed efficiently on the CPU, allowing deeper networks and larger point sets. To adapt the architecture to purely spatial neighborhoods, ArrowPose introduces two feature operators, LocalEdgeConv and SpatialEdgeConv (Eq. (2)–(3) in \cite{hagelskjaer2025arrowpose}).
LocalEdgeConv, used in the first layer, encodes only local geometric differences:
\begin{equation}
\bar{h}_{\theta}(x_j-x_i)
\label{eq:lec}
\end{equation}
Unlike EdgeConv, LocalEdgeConv omits global coordinates to emphasize local structure and mimic the locality of 2D convolutions. In subsequent layers, SpatialEdgeConv incorporates spatial distances together with learned feature differences.
To further refine the representation, we introduce two additional operators. The first, VertexConv, removes the difference term and processes only the point and neighbor features:
\begin{equation}
\bar{h}_{\theta}(x_i, x_j)
\label{eq:vc}
\end{equation}
This operator mimics standard convolution by aggregating neighbor features directly and was found to improve performance.
The second operator, EdgeVertexConv, integrates global spatial context by combining spatial differences with feature-level aggregation. Let $p_i$ and $p_j$ denote the spatial coordinates of the point and its neighbor:
\begin{equation}
\bar{h}_{\theta}(p_i, p_j-p_i, x_i, x_j)
\label{eq:evc}
\end{equation}
This operator reintroduces global spatial information at deeper layers while retaining the feature-aggregation benefits of VertexConv.
Our network uses LocalEdgeConv in the first layer, followed by SpatialEdgeConv and three VertexConv layers. Hyperparameters for all layers are provided in Table \ref{tab:network_values}. Layers 3 and 4 use a dilation factor of 2, which provides a small accuracy improvement. After down-sampling from 32,768 to 1,024 points, we apply EdgeConv with feature space neighbors; at this reduced size, the required distance matrix no longer presents memory issues. The full network architecture is illustrated in Fig.~\ref{fig:enter-label}.

\subsection{Network Training}

The network is trained using Focal Loss \cite{lin2017focal}. Focal loss is similar to cross entropy, but with a stronger weight for difficult examples. Focal loss is used as the training datasets are very unbalanced in their classes. For the main task of period classification smallest class contains only 41 samples while the largest contains 333. 
To further increase generalization we apply a jitter to the point cloud. The jitter is applied individually per point and channel as a random normal sampling. The size of the jitter is set to 3\% of the variance. The remaining training parameters are shown in Table \ref{tab:parameters}.

\begin{table}[t]
    \caption{Training parameters for the network. The same parameters are used for all tasks.}
    \label{tab:parameters}
    \centering
    \begin{tabular}{lc|lc}
    \hline
    Parameter & Value & Parameter & Value \\
    \hline
    Optimizer & SGD & Learning rate & 0.001 \\
    Scheduler & Cosine Annealing & Dropout & 0.6 \\
    Epochs & 300 &     Batch size & 10 \\
    Scheduler & Step &  Weight decay & 0.01
    \end{tabular}
\end{table}

    

\subsection{Fine-tuned Point-BERT}

We compare our method with the state-of-the-art transformer-based point cloud network, Point-BERT \cite{yu2022point}. We use the colorless version of the pre-trained weights from ULIP-2 \cite{xue2024ulip}.  
Similar to PointLLM \cite{xu2024pointllm}, we use the CLS token concatenated with the output of the MaxPool layer as input to the classification head. PointLLM used this concatenation to enable faster training, but we observed that it also improved performance. We use the same DGCNN classification head as in our method and the same training parameters, with the only difference being a larger batch size of 64. During training, we keep the Point-BERT weights frozen and only train the classification head.
Compared with our method Point-BERT uses only 8192 points as input, since the network is pre-trained with this size, and performance did not improve by increasing the number of points.

\section{Experiments}


We test our developed method and the finetuned Point-BERT on the classification tasks from \cite{bogacz2020period} and \cite{hagelskjaer2022deep}. Additionally, we introduce the new classification task and show results. Finally, we show results for an ablation study showing the contribution of the different elements. 

\subsection{Period Classification}

Bogacz et al. \cite{bogacz2020period} introduced the period-classification task for the HeiCuBeDa dataset \cite{mara2019breaking}, which contains tablets from four distinct historical periods. Because the classes are imbalanced, they limited each class to a maximum of 100 samples, resulting in 337 unique instances.
Hagelskjær \cite{hagelskjaer2022deep} instead applied a class‑balanced loss, enabling the use of a larger training set of 631 tablets. Both previous works, however, excluded tablets containing more than 2,414,753 vertices due to memory constraints.
In contrast, our method uses sub‑sampling and is therefore independent of the original CAD model resolution. This allows us to include all tablets, creating a full training set of 747 instances, with 121, 41, 252 and 333 instances in the four different classes. For consistency with other methods, all experiments use the same test data.
Table \ref{tab:period} reports the period‑classification results as the average F1‑score across classes. On the small dataset, the pre‑trained Point‑BERT backbone with a fine‑tuned MLP head outperforms previous methods. On the medium dataset, its performance is slightly lower, but when all samples are included, it again improves.
Across all dataset sizes, our proposed method achieves the highest performance. Notably, it surpasses all baselines even when trained only on the small dataset. Using the full dataset of 747 tablets, it reaches a new state‑of‑the‑art score of 99\%.

\begin{table}[t]
\caption{Average F1-score for period classification with the three different dataset sizes. }
\label{tab:period}
\centering
    \begin{tabular}{lccc}
    \hline
    Training dataset size                                   & 337  &  631 & 747 \\ 
    \hline
    Bogacz \cite{bogacz2020period}                          & 0.84 &      &       \\
    Hagelskjær \cite{hagelskjaer2022deep}                   & 0.88 & 0.92 &       \\
    Point-BERT                                               & 0.89 & 0.91 & 0.93  \\  
    Ours                                                    & 0.96 & 0.97 & \textbf{0.99} 
    \end{tabular}
\end{table}


\begin{figure}[t]
    \centering
    \includegraphics[width=0.90\linewidth]{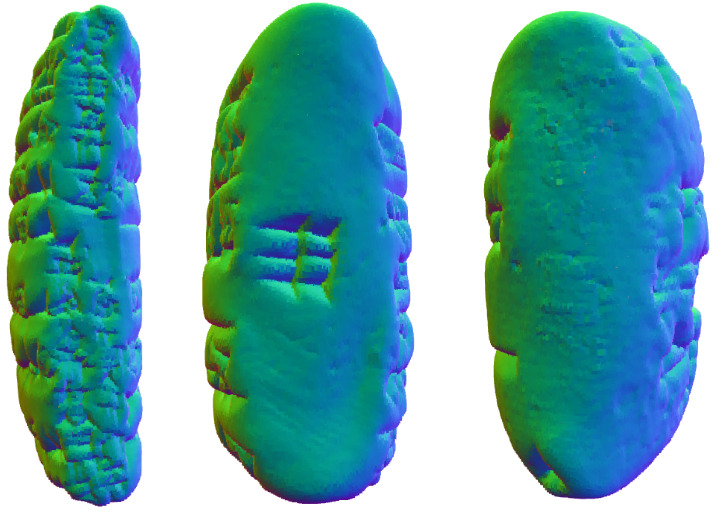}
    \caption{Three tablets shown from the left side ("HS 1238", "HS 2271", "HS 2274"). The larger curvature on the left side is seen, except for tablet "HS 2274" which is wrongly oriented. On the two leftmost tablets the left side sign is present.}
    \label{fig:obverse}
\end{figure}

\subsection{Seal and Left Side Sign}


We also evaluate our method on the two classification tasks introduced in \cite{hagelskjaer2022deep}: detecting whether a seal is present on the tablet and determining whether written signs appear on its left side. Examples of the left side sign are shown in Fig.~\ref{fig:obverse}. The results for both tasks are reported in Table \ref{tab:sealleft}. Point‑BERT achieves performance comparable to the method presented in \cite{hagelskjaer2022deep}. However, our method outperforms both baselines on all datasets and reaches 100\% accuracy on the seal‑presence task. This demonstrates that our network is also effective on tasks that require attention to local regions of the tablet.

\begin{table}[t]
    \caption{Average precision for meta data classification tasks.}
    \label{tab:sealleft}
    \centering
        \begin{tabular}{lcc}
        \hline
        Method          & Seal presence & Left side sign \\ 
        \hline
        Hagelskjær \cite{hagelskjaer2022deep}      & 0.91      & 0.94  \\
        Point-BERT                                  & 0.91      &  0.91  \\
        Ours                                        & 1.0      & 0.97 
        \end{tabular}
\end{table}


\subsection{Tablet Front}

In this paper we also introduce an additional metadata classification task. This task is about classifying if the front of the tablet is facing towards or away from the camera. This is a challenging task as outlined in the paper presenting the dataset 
\cite[page 5]{mara2019breaking}: \begin{quote}
This work was done by an Archaeologist to ensure a reasonably large number of correctly orientated tablets. 
At most, we expect front- and backsides being confused...
\end{quote}
However, the 3D shape of the object can help to determine the orientation of the tablet as the front side is generally flatter. We generate the dataset using the HeiCuBeDa dataset \cite{mara2019breaking}. Similar to \cite{hagelskjaer2022deep} we limit the dataset to only use tablet from the  Ur III time period. We then remove all tablets which does not have writing on both the front and backside, to ensure they look similar on both sides. Positive samples are generated by simply using the point cloud, and negative samples are generated by rotating the point 180 degrees around the x-axis. The tablets are rotated about the x-axis as scribes flipped it this way when moving to the backside. 
The dataset is split 90/10 for training and testing, resulting in 319 and 35 samples.  
The accuracy of the Point-BERT method on the dataset is 77\% and for our method the accuracy is 98.5\%.   
If we require that both views of each tablet should agree to make a prediction we get a precision of 100\% and a coverage of 97\%.
There was one wrong prediction, "HS 2274", but when comparing with the image from the CDLI (Cuneiform Digital Library Initiative) database \cite{CDLI2026Home} it appears wrongly oriented in the HeiCuBeDa dataset. 
The good performance of our method demonstrate that deep learning could be used for orientating scans of tablets, while difficult cases could be handed over to archaeologists.
The wrongly oriented tablet and other examples are shown in Fig.~\ref{fig:obverse}.

\subsection{Ablation}


In this ablation study, we evaluate the contribution of each component to the overall network performance. We report results for the period-classification task using both the small and large training sets, as shown in Table \ref{tab:ablation}. The experiments indicate that normal vectors provide the largest performance gain, while dilation contributes the least.

As Point-BERT only uses 8192 points we have also compared with our method when equalizing the number of points. The results are shown in Table \ref{tab:nrpoints}. 
Although the performance of our method decreases when the point cloud resolution is reduced, it still consistently outperforms Point‑BERT. Conversely, increasing the number of input points leads to a performance drop for Point‑BERT.
This flexibility to the number of points is expected as our method is not pretrained on a fixed number of points.

\begin{table}[t]
\caption{Resulting F1-score for period classification in ablation study.}
\label{tab:ablation}
\centering
    \begin{tabular}{lcc}
    \hline
    Omitted      & 337 & 747 \\ 
    \hline
    Dilation     & 0.96 &  0.97      \\
    Normal Vectors & 0.90 & 0.95     \\
    VertexConv    & 0.93 & 0.96       \\
    1D-Conv       &  0.93  &  0.96         \\
    DGCNN Layer  &  0.93 &  0.96     \\
    None         & 0.96 & 0.99       
    \end{tabular}
\end{table}

\begin{table}[t]
    \caption{Resulting F1-score for period classification with varying number of points in the point cloud.}
    \label{tab:nrpoints}
    \centering
    \begin{tabular}{lcccc}
        \hline
         Training dataset size      & 337 & 747 & 337 & 747 \\
         Input point cloud size     & 8192 & 8192 & 32,768 & 32,768 \\
         \hline
         Point-BERT                 & 0.89 & 0.93 & 0.84 & 0.93  \\
         Ours                       & 0.93 & 0.96 & 0.96 & 0.99
    \end{tabular}
\end{table}

\section{Conclusion}

This paper introduced a novel network architecture for processing large point clouds and demonstrated its effectiveness across multiple datasets for cuneiform tablet metadata classification. On all benchmarks, our method achieves state-of-the-art performance and consistently outperforms the transformer-based Point-BERT baseline. We also proposed a new classification task focused on detecting wrongly oriented tablets, where our model performed strongly and successfully identified a mislabeled sample in the dataset. These results highlight the potential of our approach as a robust, framework for metadata classification.

In future work, the method can be evaluated on additional 3D tablet datasets or extended to entirely different point cloud tasks with similar structural characteristics. Since we used a pre-trained version of Point-BERT as a baseline, another avenue is to retrain Point-BERT from scratch using its full training scheme to better match the domain of cuneiform tablets. Finally, combining our method with recent LLM-based point cloud models for text prediction, such as PointLLM \cite{xu2024pointllm}, may further enhance the ability to link 3D geometry with semantic metadata or translation.

\bibliographystyle{IEEEtran}
\bibliography{references}

\end{document}